\documentclass{article}

\PassOptionsToPackage{numbers, compress}{natbib}

\usepackage[preprint]{neurips_2019}

\usepackage[utf8]{inputenc}   
\usepackage[T1]{fontenc}      
\usepackage{hyperref}         
\usepackage{url}              
\usepackage{booktabs}         
\usepackage{amsfonts}         
\usepackage{nicefrac}         
\usepackage{microtype}        
\usepackage{graphicx}
\usepackage{subcaption} 
\usepackage{amstext}
\usepackage{bm}
\usepackage{enumitem}
\usepackage{amsthm}
\newtheorem{proposition}{Proposition} 
\newtheorem{remark}{Remark}  

\title{A Polynomial-Based Approach for Architectural Design and Learning with Deep Neural Networks}

\author{%
  Joseph ~Daws~Jr., \thanks{website: joedaws.github.io} \\
  Department of Mathematics\\
  University of Tennessee at Knoxville\\
  Knoxville, TN 37996 \\
  \texttt{jdaws@vols.utk.edu} \\
   \And
   Clayton~G. Webster \\
   Computational and Applied Mathematics\\
   Oak Ridge National Lab\\
   Oak Ridge, TN 37831    \\
   \texttt{webstercg@ornl.gov}, \\
   and \\
  Department of Mathematics\\
  University of Tennessee at Knoxville\\
  Knoxville, TN 37996 \\
  \texttt{cwebst13@utk.edu}
}

\begin{document}

\maketitle

\begin{abstract}
    In this effort we propose a novel approach for reconstructing multivariate
    functions from training data, by identifying both a suitable network
    architecture and an initialization using polynomial-based approximations.
    Training deep neural networks using gradient descent can be interpreted
    as moving the set of network parameters along the loss landscape in order to
    minimize the loss functional. The initialization of parameters is important for
    iterative training methods based on descent. Our procedure produces a network 
    whose initial state is a polynomial representation of the training data. 
    The major advantage of this technique is from this initialized state the 
    network may be improved using standard training procedures. 
    Since the network already approximates the data,
    training is more likely to produce a set of parameters associated with a
    desirable local minimum. We provide the details of the theory necessary for 
    constructing such networks and also consider  
    several numerical examples that reveal our approach ultimately produces
    networks which can be effectively trained from our initialized state to achieve an improved
    approximation for a large class of target functions.
\end{abstract}

\section{Introduction}

Deep neural networks (DNNs) have emerged as powerful nonlinear approximation tools and
have been deployed with great success in many challenging tasks such as 
image classification \cite{NIPS2012_4824}, 
playing games, such as Go, at a world-class level \cite{alpha_go},
and even to produce examples which fool other classifiers \cite{NIPS2014_5423}. 
However, DNNs are also known to be difficult to train \cite{glorot2010understanding}. 
Each deep network has its own set of hyper-parameters which 
must be defined, e.g., the number of layers, the number of nodes per layer,  
and the connectivity between the nodes on different layers. 
Ideally, the choice of these parameters is  
made with respect to the available training data and the task to be 
solved by the network. In this paper we identify suitable deep network 
architectures, based on polynomials.
Gradient descent-based training procedures are known to be effective for identifying
good network parameters \cite{bottou2010large}.
However, such algorithms are sensitive to the initial set of parameters.
In addition to identifying suitable network architectures, based on training data, we
provide an initialization of parameters so that they perform at least as well as 
a given polynomial approximation of the training data.
This paper establishes an explicit relationship between polynomial approximation 
and approximation by a neural network and shows that certain network architectures 
can perform at least as well as any given polynomial-based approach. 
We also provide numerical examples showing our network not only produces a good 
approximation but also that our initialization makes training more efficient. 

From a high level perspective, every task that a neural network solves can be 
characterized as a function approximation problem.
For example, consider the classification of the ImageNet data 
set \cite{imagenet_cvpr09}, which is composed of many images which fall into 
one of many possible classes. 
A typical approach to solve this problem is to choose $k$ classes of images and to
train a classifier whose input is an image and whose output is 
a vector of $k$ probabilities. Each component of the output vector represents 
the probability that the input belongs to the class
corresponding to the component. Therefore, the task is solved by finding a suitable 
function from a very high dimensional space to a $k$-dimensional space. 
Classical approximations, that utilize a basis or frame,
have a long and very successful history in many diverse 
areas of science. As such, by constructing networks which achieve comparable
performance to polynomial approximations we can explore how neural network approximations 
relate to other forms of classical, but highly non-linear approximation, such as 
$n$-term approximations, dictionary approaches, etc. 

As Neural networks begin to be integrated into fault intolerant 
real-world systems, such as self driving cars \cite{selfdrivingcar}, 
understanding the error between the network and 
the desired task/function is vital. 
Moreover, since it is known that neural networks are universal 
approximators \cite{Cybenko1989ApproximationBS,NIPS2017_7203}, 
it is clear that one can build and train an arbitrarily accurate network given 
enough samples of the target function and given the 
ability to construct a network as large as desired.
There has been extensive research into constructing approximations
by classical functions and, in particular, polynomials, see, e.g., 
\cite{rons_good_book}. Moreover, these constructive approximations 
have sharp convergence rates associated with
their errors in a variety of norms. Such results may be helpful for creating 
neural networks which obtain high fidelity approximation of a target function. 
Therefore, by initializing a network to have the 
same behavior as a high-fidelity approximating polynomial, we not only have a network
whose error is well understood, 
but also can further train the network to \textbf{possibly achieve an even 
more accurate approximation}. 

In what follows, we propose a network architecture with a sufficient
number of nodes and layers so that 
it can express much more complicated functions than the polynomials used to 
initialize it. In Section \ref{poly_net} we outline the construction
of two networks which approximate polynomials. The first can approximate a
given polynomial. Numerical examples
which show the performance of this network for approximating a target
function are given in Section \ref{numerical_section} where
we initialize a network to a polynomial
approximation of the training set and then train it 
to achieve better performance. The architecture of the first network constructed in
Section \ref{poly_net} is not necessarily the same as those used in practice. 
It is composed of many simple but separate sub-networks. However,
the second network we construct, associated with a specific polynomial,
is a deep, feed-forward network with the same number of nodes on each of its
hidden layers. Such an architecture is widely used and in Section
\ref{numerical_section} we show that our polynomial-based initialization
allows for easier training and better performance for approximating a
given target function.

\subsection*{Related Work}
Several other efforts have considered constructing networks which achieve polynomial
behavior \cite{deep_relu_du,deep_learning_schwab,Yarotsky_2017} wherein networks
are constructed that approximate polynomials associated with sparse grids, 
Taylor polynoimals and generalized polynomial chaos approximations. 
The network presented in this paper is a slightly modified one presented in 
\cite{deep_learning_schwab}. Those authors constructed a network which approximates the
product of $n$ inputs and used this network to compute multivariate Taylor polynomials.
Choosing suitable initialization of network parameters was considered in 
\cite{thimm1995neural}. A random initialization scheme which avoids common training 
failures was presented in \cite{hanin2018start}.

\section{A Network which Approximates a Polynomial}
\label{poly_net}
In this section, we construct a network which approximates a given
polynomial arbitrarily well and consider a specific example which is 
implementable by a deep, feed-forward architecture. 
Our network will be able to approximate $d$-dimensional
polynomials of the form 
	\begin{equation} \label{eq:poly_fun}
		F(\vec{x}) = \sum_{\vec{\nu} \in \Lambda} c_{\vec{\nu}} 
		\Psi_{\vec{\nu}} (\vec{x})
	\end{equation}
where $\vec{x} \in \mathbb{R}^d$, $\vec{\nu} \in \mathbb{N}_0^d$ is a multi-index,
$\Lambda$ is a finite set of cardinality $N$, and 
$c_{\vec{\nu}}$ is the coefficient associated with the polynomial
$\Psi_{\vec{\nu}}$,
which is a tensor product of one-dimensional polynomials.
Each polynomial in the sum in (\ref{eq:poly_fun}) is assumed to be of the form,
	\begin{equation} \label{eq:tensor_poly}
		\Psi_{\vec{\nu}}(\vec{x}) = \prod_{i=1}^d \psi_{\nu_i}(x_i)
	\end{equation}
where $\vec{\nu} = (\nu_{1},\dots,\nu_{d})$, and $\psi_{\nu_i}$ 
is a single variable polynomial of degree $\nu_i \in \mathbb{N}_0$.
The polynomial in (\ref{eq:tensor_poly}) can be computed by finding the product
of $d$ numbers, and, by the fundamental theorem 
of algebra, $\psi_{\nu_i}(x_i)$ can be computed by a product of
$\nu_i$ possibly complex numbers. That is,
    \begin{equation} \label{eq:fun_thm_alg}
        \psi_{\nu_i}(x_i) = a_i\prod_{k=1}^{\nu_i} 
        \left( x_i - r_k^{(\nu_i)} \right)
    \end{equation}
where the numbers $r_k^{(\nu_i)} \in \mathbb{C}$ are called 
the roots of the polynomial
$\psi_{\nu_i}(x_i)$ and $a_i$ is a scaling factor. Hence, 
    \begin{equation} \label{eq:prod_all_roots}
        \Psi_{\vec{\nu}}(\vec{x}) = \prod_{i=1}^d a_i \prod_{k=1}^{\nu_i} 
        \left(x_i - r_k^{(\nu_i)} \right).
    \end{equation}
Without loss of generality, 
we will focus on the case when the polynomials $\psi_{\nu_i}$ have real roots,
which is a reasonable restriction since all orthogonal, 
univariate polynomials have real roots.
Moreover, these polynomials can be used to form a basis for polynomials with 
complex roots in which the polynomial with complex roots can be represented exactly.

\begin{figure}
    \begin{subfigure}{.5\textwidth}
      \centering
      \includegraphics[width=0.8\linewidth]{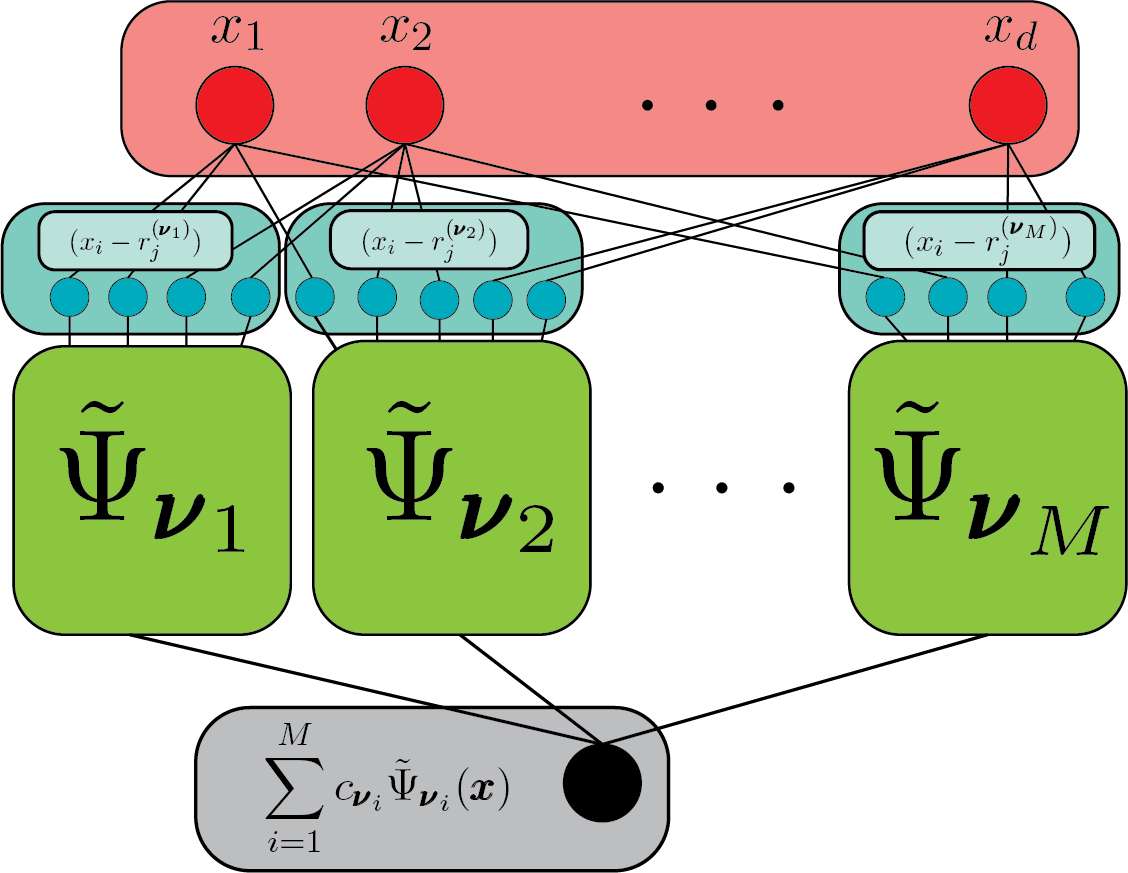}
      \caption{}
      \label{fig:F_tilde}
    \end{subfigure}%
    \begin{subfigure}{.5\textwidth}
        \centering
        \includegraphics[width=0.8\linewidth]{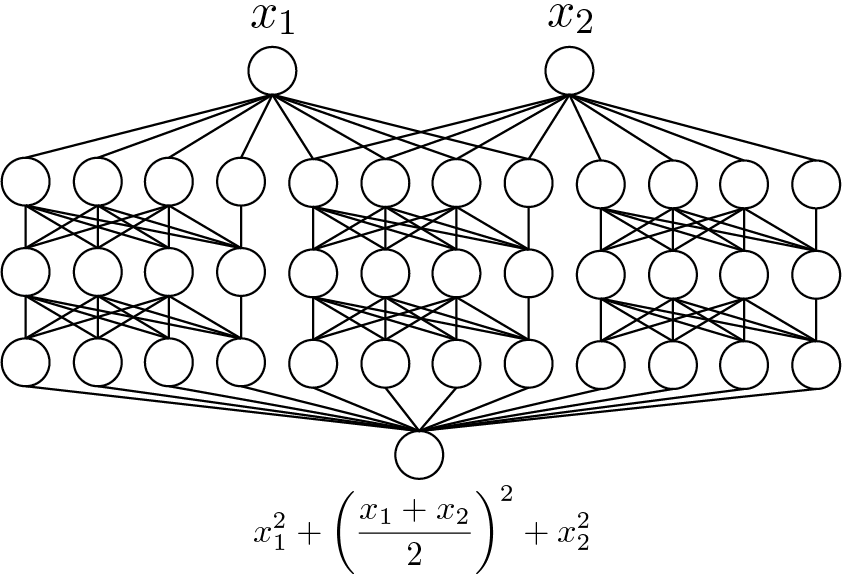}
        \caption{}
        \label{fig:x1x2_net}
    \end{subfigure}%
    \caption{ (a) An outline of a network which computes an approximation of the polynomial
    $F$ as in (\ref{eq:poly_fun}). The first layer contains nodes which compute the
    shift of the inputs $x_i$ by the appropriate root $r_k^{(\nu_i)}$ as in Step 4 of our construction procedure. 
    Step 5 is accomplished by passing the outouts to a sub-network $\tilde{\Psi_{\vec{\nu}}}$ 
    which computes the product of the roots as in (\ref{eq:prod_all_roots}). Finally, Step 6 
    is accomplished by a linear combination
    of the outputs of the sub-networks $\tilde{\Psi_{\vec{\nu}_i}}$ so that it 
    approximates the polynomial (\ref{eq:poly_fun}).
    (b) A network which approximates the polynomial $\tilde{S}$ where $d=2$. This network 
    can also approximate the product $x_1 x_2$ using a suitable linear combination of the outputs
    $x_1^2$, $x_2^2$ and $((x_1 + x)/2)^2$.}
    \label{fig:polynomial network}
\end{figure}

Before going into the details we outline the general construction of a network $\tilde{F}$ which approximates 
$F$.
An illustration of its structure is depicted in Figure \ref{fig:polynomial network} (b).
    \begin{enumerate}[label=Step \arabic*:]
        \item Choose a family of univariate polynomials $\{ \phi_i \}_{i=1}^\infty$, 
              such as the Legendre polynomials.
        \item Find an approximation of the training 
        data $\{(\vec{x}_j,F(\vec{x}_j)\}_{j=1}^M$ in the tensor product basis
        generated by $\{ \psi_i \}$ with $N$-terms associated with a chosen 
        index set $\Lambda$ by identifying the coefficients $c_{\vec{\nu}}$. 
        \item Find the necessary roots $r_k^{(\nu_i)}$ from (\ref{eq:prod_all_roots}) and values $a_i$.
        \item The first layer of the network takes the inputs $\vec{x}$ and 
        sends them to a linear layer with $\sum_{\nu \in \Lambda} \| \vec{\nu} \|_1$
        nodes each of which computes $x_i - r^{(\nu_i)}$.
        \item These values are then used as inputs to sub-networks 
        $\tilde{\Psi}_{\vec{\nu}_i}$ which
        approximate the product polynomial $\Psi$ given by (\ref{eq:tensor_poly}).
        \item The output weights are set to be the coefficients $c_{\vec{\nu}}$ and 
        the product polynomial is then approximated by the linear combination of the
        outputs of each of the product blocks $\tilde{\Psi}_{\vec{\nu}_i}$. 
    \end{enumerate}

In light of (\ref{eq:prod_all_roots}), 
approximating the polynomial $\Psi_{\vec{\nu}_j} (\vec{x})$ 
can be accomplished by constructing a network that computes the product of 
$\|\vec{\nu}\|_1 := \nu_1 + \nu_2 + \cdots + \nu_d$ numbers. 
Such a network can constructed by first constructing a network which computes the 
product of $2$ numbers. Copies of these sub-networks can be chained together $n-1$
times to produce the product of $n$ numbers. 
This approach was used in \cite{deep_learning_schwab} where a series of smaller sub-networks 
$\tilde{p}(x_i,x_j)$ which approximate the product $x_i x_j$ were arranged into a binary 
tree in order to compute the product of $n$ numbers. The network $\tilde{p}$
is constructed by noting that 
    \[
        x_i x_j = \frac{1}{2} \left ( (x_i+x_j)^2 - x_i^2 - x_j^2 \right )
    \]
so that the desired product can be approximated by a linear combination of networks 
which approximate $x \mapsto x^2$.
A network for such a mapping has been constructed in 
\cite{liang2016deep,Yarotsky_2017,deep_learning_schwab} but only for inputs on the 
interval $[0,1]$. 

\begin{figure}
    \begin{subfigure}{.33\textwidth}
        \centering
        \includegraphics[scale=0.27]{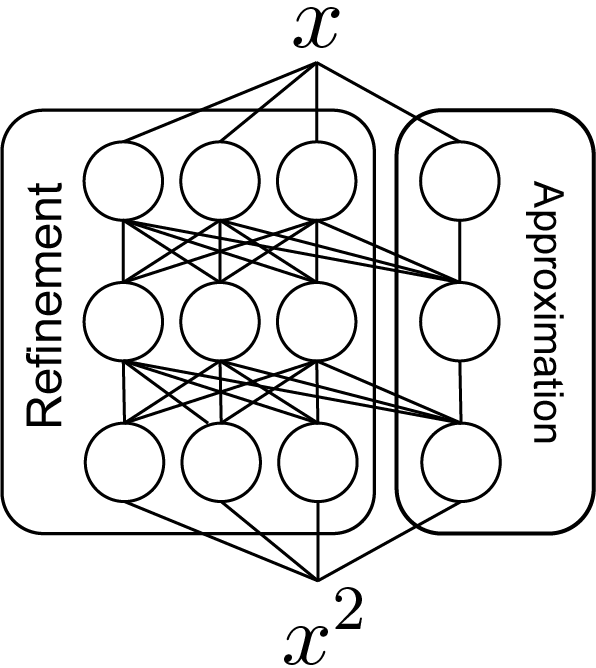}
        \caption{}
        \label{fig:x2_net}
    \end{subfigure}%
    \begin{subfigure}{.33\textwidth}
      \centering
      \includegraphics[width=\linewidth]{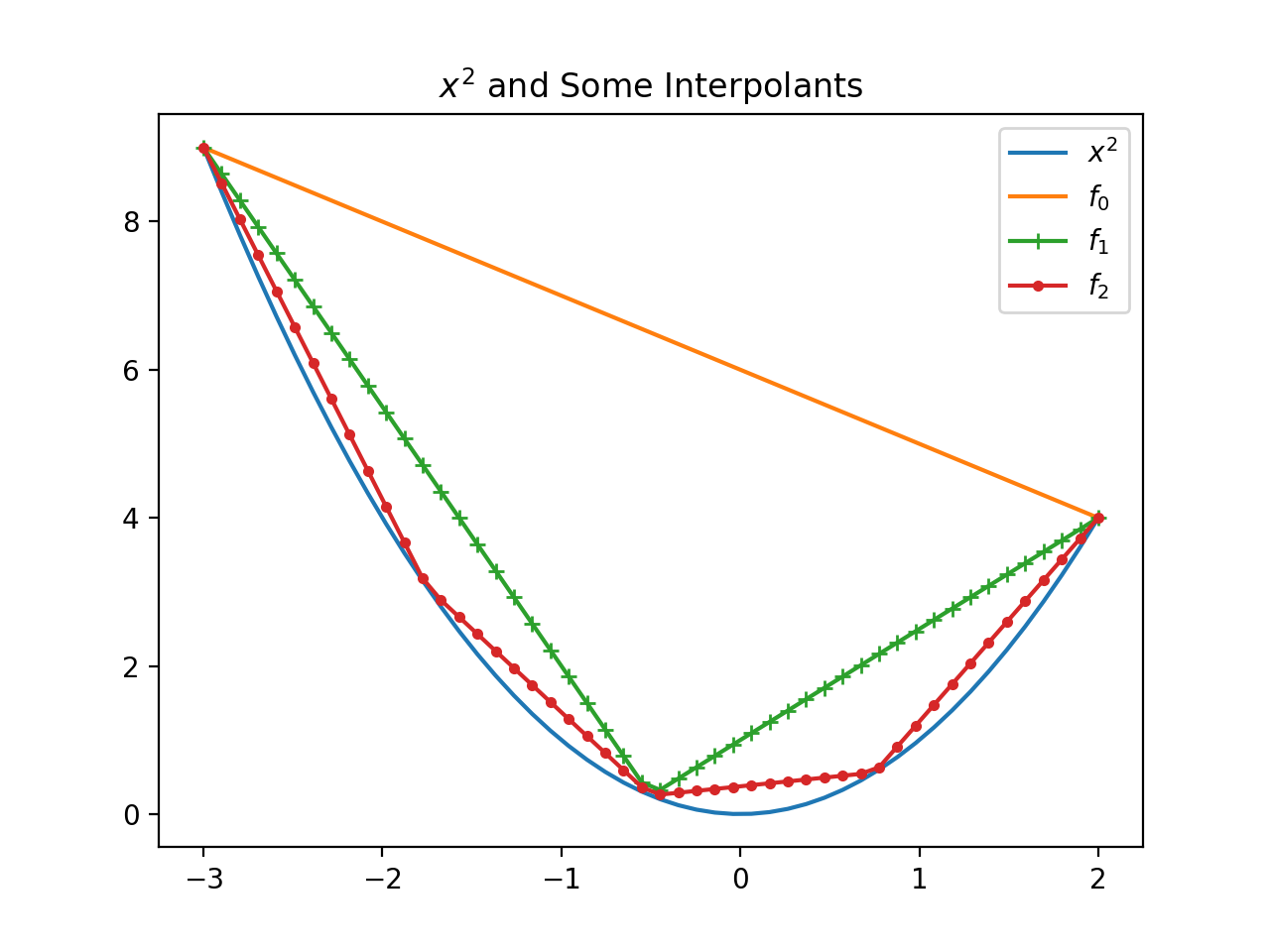}
      \caption{}
      \label{fig:x2_interps}
    \end{subfigure}
    \begin{subfigure}{.33\textwidth}
      \centering
      \includegraphics[width=\linewidth]{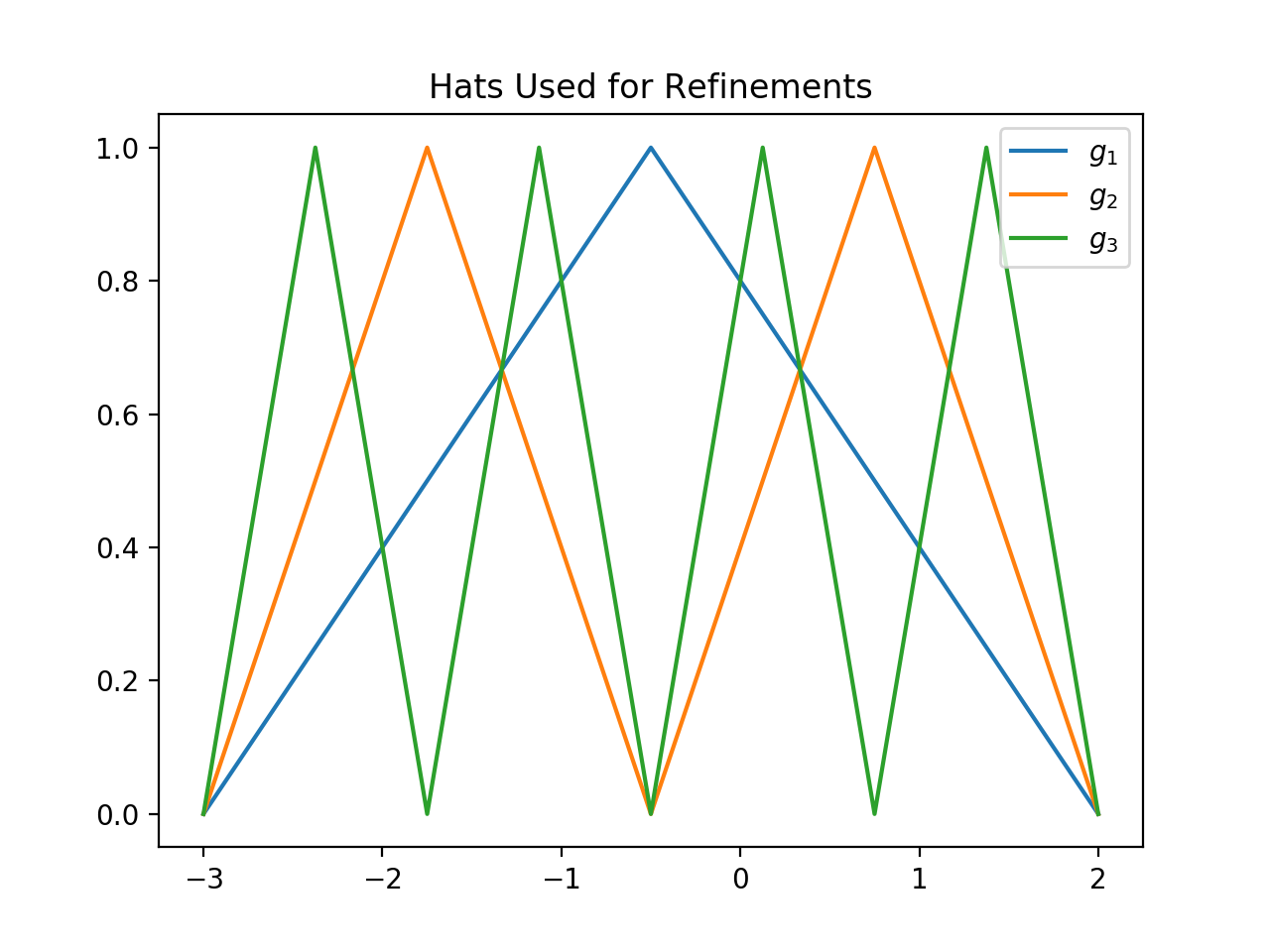}
      \caption{}
      \label{fig:hats}
    \end{subfigure}
    \caption{ (a) A network architecture for approximating $x^2$ on any interval [a,b].
               (b) Example of the piecewise linear interpolants 
               on the interval $[-3,2]$ used in the
               approximation part of the network.
               (c) Example of the hat functions used to refine the piecewise linear 
               interpolants on the interval $[-3,2]$.}
    \label{fig:x2_network_ab}
\end{figure}

These networks are not suitable for computing
the product polynomial given by \ref{eq:prod_all_roots}. Assume that $\vec{x} \in [a,b]^d$. 
Recall that if the $\phi_{\nu_i}$'s are orthonormal, univariate polynomials on $[a,b]$
then their roots are real, in the interval $[a,b]$, and we have that
$2a \le \left| x_i - r_k^{(\nu_i)}\right| \le 2b$. In order to compute
a polynomial (\ref{eq:prod_all_roots}), we need to ensure that 
the product of numbers in the interval 
$\left[ (2a)^{|\vec{\nu}_j|},(2b)^{|\vec{\nu}_j|} \right]$
can be computed. However, the same network architecture, as depicted in 
Figure \ref{fig:x2_network_ab} (a), used in \cite{Yarotsky_2017,deep_learning_schwab} 
can be used to approximate $x^2$ on a general interval $[a,b]$ by changing 
the network parameters. The following proposition establishes the existence
of sucha network and gives explicit parameters so that the network can be constructed.

\begin{proposition} \label{prop:x2_ab_net}
    A network $\tilde{f}$ with $L$ hidden layers and $4$ nodes on 
    each layer can approximate
    the function $f(x) = x^2$ on the interval $[a,b]$ 
    with
        \[
            \sup_{x \in [a,b]} | x^2 - \tilde{f} | \le \frac{C}{2^{2L}},
        \]
    where $C = (b-a)^2/4$.
\end{proposition}

\begin{proof}
    Let $f_m$ be the piecewise linear
    interpolant of $x^2$ on $[a,b]$ so that for
    $\xi_{k,m} := a + k(b-a)/2^m$ where $k=0,\dots,2^m$
    we have $f_m(\xi_{k,m}) = f(\xi_{k,m})$. The functions $f_0$, $f_1$, and $f_2$
    are plotted in Figure \ref{fig:x2_network_ab} (b) for $x^2$ on the interval
    $[-3,2]$. The proof of this proposition has two parts. First, we will show
    that $f_m$ can be represented as a linear combination of the composition of 
    some special functions. Then, we will show that these functions 
    can be implemented by a linear combination of ReLU functions with specifically
    chosen weights and biases. The desired network uses these parameters to compute
    $f_m(x)$. The error can be computed using a standard error estimate for 
    piecewise linear interpolants. 
    
    Notice that 
    \[
        f_{0}(\xi_{k,m}) - f_{1}(\xi_{k,m}) =
        \left \{ 
        \begin{array}{cc}
             0  & \;\; \text{ if $k$ is even } \\
             C & \text{ if $k$ is odd}, 
        \end{array}
        \right. \nonumber
    \]
    where $C := (b-a)^2/4$. Since $f_{0}(x) - f_{1}(x)$
    must be linear on each of the intervals $[a,(a+b)/2)$ and $[(a+b)/2,b]$,
    we may write $f_{0}(x) - f_{1}(x) = C g_1(x)$ where 
    \[
        g_1(x) = \left \{ 
            \begin{array}{cc}
                \frac{2}{b-a} (x - a) & \; a \le x < \frac{a+b}{2} \\
                \frac{2}{a-b} (x - b) & \; \frac{a+b}{2} \le x \le b.
            \end{array}
        \right.
    \]
    We can derive a similar equation for each of the differences
    $f_{m-1}(x) - f_m(x)$. Let $h(x)$ the ``hat" function on the interval
    $[0,1]$ given by
    \[
        h(x) = \left \{
            \begin{array}{cc}
                 2 x & 0 \le x < 1/2  \\
                 2(1-x) & 1/2 \le x \le 1.
            \end{array}
        \right .
    \]
    Let $g_2(x) := h(g_1(x))$.
    Since the range of range of $g_1$ is $[0,1]$, it
    achieves each of its values twice, and has an axis of 
    symmetry about $x = (a+b)/2$, the the composition $h(g_1(x))$
    is the two-hat function so that 
    \[
        g_2(\xi_{k,2}) = \left \{ 
            \begin{array}{cc}
                0 & \text{ if $k$ is even}  \\
                1 & \text{ if $k$ is odd} 
            \end{array}
        \right. 
    \]
    For example, $g_2$ for the interval $[-3,2]$ is plotted in Figure 
    \ref{fig:x2_network_ab} (c).
    Now we notice that $f_1(x) - f_2(x) = \frac{C g_2(x)}{2^2}$. For a general $m$ 
    we have
    \begin{equation} \label{eq:chain_equation}
        f_{m-1}(x) - f_m(x) = \frac{C g_m(x)}{2^{2m}}.
    \end{equation}
    where $g_m(x) := h(h(\cdots h(g_1(x))\dots))$ is the function defined by
    $h$  applied to the output of $g_1(x)$ $m-1$ times. 
    An equation for $f_m$ can now be derived by sequentially applying 
    (\ref{eq:chain_equation}),
        \begin{equation} \label{eq:interp_eq}
            f_m(x) = f_0(x) - \sum_{i=1^m} \frac{C g_i}{2^{2m}}.
        \end{equation}
    
    Our network $\tilde{f}$ is constructed so that its output is $f_m(x)$, that is,
    $\tilde{f}(x) = f_m(x)$ for all $x \in [a,b]$. It is possible to
    express $f_0(x)$ on the interval $[a,b]$ as a single ReLU function, i.e.,
    $f_0(x) = \sigma \left( (a+b) x - ab \right )$.
    Both $g_1(x)$ and $h(x)$ can be written as linear combinations of $3$ ReLU
    nodes. We have $g_1(x) = (2/(b-a))\sigma(x-a) + (4/(b-a))\sigma(x-(a+b)/2) 
    + (2/(b-a))\sigma(x-b)$ 
    and $h(x) = 2\sigma(x) - 4\sigma(x-1/2) + 2\sigma(x)$. 
    Then according to (\ref{eq:interp_eq}), $f_m(x)$ can be computed by a network with 
    $m$ hidden layers and $4$ nodes on each layer. This can be seen in 
    Figure \ref{fig:x2_network_ab} (a). The left three nodes on each layer labeled 
    ``refinement" are used to compute either 
    $g_1$ or $h$ and the nodes labeled ``approximation" are used to compute $f_k$
    for $k=0,\dots,m-1$. The output node computes $f_m(x)$.
    
    Since $\tilde{f}$ is the piecewise interpolant 
    $f_m$ of $x^2$, $\sup_{x \in [a,b]} 
    | x^2 - \tilde{f} | <\frac{C}{2^{2m}}$.
\end{proof}

Using the network $\tilde{f}$ to approximate 
$x^2$ on the interval $[a,b]$ we can construct a network
which approximates the product $x_i x_j$ for $x_i,x_j \in [a,b]$ 
by taking a linear combination of
$x_i^2$, $(x_i+x_j)^2$, and $x_j^2$.
In order to accurately approximate $(x_i+x_j)^2$ with $\tilde{f}$ we 
consider the following network
    \begin{equation} \label{ab_net}
        \tilde{\times}(x_i,x_j) 
        = \frac{1}{2} \left ( 4 \tilde{f} \left(\frac{x_i+x_j}{2} \right) 
        - \tilde{f}(x_i) - \tilde{f}(x_j)
        \right).
    \end{equation}
We must scale the quantity $x_i + x_j$ so that it is in the interval [a,b]
before applying the network $\tilde{f}$.
This network has similar error rate and complexity as those constructed in 
\cite{Yarotsky_2017, deep_learning_schwab} up to a slightly different constant 
which depends on $a$ and $b$ and with one less layer, since we do not require 
the absolute values of $x_i$, $x_j$ and $x_i+x_j$. 
The absolute value of $x$ can be computed using ReLU activation functions 
by noticing that $x = \sigma(x) - \sigma(-x)$. However, to compute this would require
an additional ReLU layer. Since $\tilde{f}$ constructed in the previous 
proposition can approximate $x^2$ on any interval $[a,b]$, 
our network does not require us to find the absolute values
of the inputs.

A network $\tilde{\Psi}_{\vec{\nu}}(\vec{x})$ which approximates 
$\Psi_{\vec{\nu}}(\vec{x})$ can be constructed as a sequence of 
compositions of copies of a given sub-networks $\tilde{\times}$ which approximate the product 
of two numbers. That is,
    \begin{equation} \label{eq:the_net}
        \tilde{\Psi}_{\vec{\nu}}(\vec{x}) := 
        \tilde{\times}(A,\tilde{\times}(w_{\|\vec{\nu}_j\|_1},\tilde{\times}(\dots,\tilde{\times}(w_3,\tilde{\times}(w_1,w_2))))),
    \end{equation}
where the value $A:= \prod_{i=1}^d a_i$ that can be computed 
once all of the univariate polynomials
$\psi_{\nu_{i}}$ are fixed and where $\{ w_k \}$ is an enumeration
of linear combination of the inputs and a root, 
i.e., $x_i - r_k^{(\nu_i)}$. 
Using $N$ networks like (\ref{eq:the_net}), we can approximate the polynomial
$F_{\Lambda}$ by the network
    \begin{equation}
        \tilde{F}_{\Lambda}(\vec{x}) := \sum_{j=1}^N c_{\vec{\nu}_j} \tilde{\Psi}_{\vec{\nu}_j}(\vec{x}).
    \end{equation}
Figure \ref{fig:polynomial network} (a) outlines the structure of our network.
In Section \ref{numerical_section}, our numerical experiments consider initializing 
a network with the structure of $\tilde{F}$ and then training it subject to a set of training data.
Our first numerical example in Section \ref{numerical_section} explores using the architecture depicted 
in \ref{fig:polynomial network} to approximate a rational function.

The architecture of the network $\tilde{F}$ is somewhat unrealistic. It is primarily composed
of small sub-networks and does not allow connections between interior nodes in separate sub-networks. 
We now propose a network and initialization which has a realistic architecture, 
i.e., a deep, fully connected, feed-forward neural network with the same number of nodes on each 
of the hidden layers. Our network can be initialized to approximate the polynomial
    \begin{equation} \label{eq:proposed_net}
        \tilde{S}(\vec{x}) := \sum_{i=1}^{d} x_i^2 + \sum_{i=1}^{d-1} \frac{1}{4} (x_i + x_{i+1})^2.
    \end{equation}
The 2-dimensional case for $\tilde{S}$ is given in Figure \ref{fig:polynomial network} (b) 
which is composed of three $\tilde{f}$ networks in parallel. The middle four nodes on each layer 
are associated with $\tilde{f}((x_1 + x_2)/2)$. For the $d$-dimensional case,
$\tilde{S}$ is a deep, full connected neural network with $d$ input units,
$L$ hidden layers with $4*(2d-1)$ nodes on each layer, 
and $1$ output unit. Although we initialize the parameters 
of $\tilde{S}$ so that each instance of $\tilde{f}$ is not connected, once training begins, 
non-zero weights between any two nodes on consecutive layers may form. Thus $\tilde{S}$ provides an
initialization for a deep, fully connected network. We give several numerical 
examples in Section \ref{numerical_section} which show that this architecture and 
initialization is effective for learning complicated functions. Moreover, our 
numerics suggest that our initialization may prevent over-fitting of the training data.

\begin{remark}
We have constructed the network $\tilde{F}$ to approximate 
a given polynomial $F$. One can view $F$ as a hyper-parameter of the network 
$\tilde{F}$ since it identifies a specific architecture as well as a set of initial values. 
We will briefly discuss how one might chose a suitable approximating polynomial
given a set of training data. In polynomial approximation it is common to 
put some assumptions on the target function. For instance, many theorems make 
assumptions on the smoothness of the function, the distribution of the sample points, 
or the sparsity in a certain polynomial basis.
This information can be used to choose an appropriate 
polynomial approximation scheme which is then used to both generate a network 
and initialize its values in order to achieve comparable error. 
Finally, we can perform more training on the network so that it 
achieves an approximation that is better than the polynomial used to initialize it.
\end{remark}

\section{Numerical Examples}
\label{numerical_section}
All of the numerical experiments below were implemented
using \texttt{PyTorch}. For training, we use the
mean square error loss functional 
$\frac{1}{n} \sum_{i=1}^n (F(\vec{x}_i) - \tilde{F}(\vec{x}_i))^2$
and the the ADAM optimizer proposed in \cite{KingmaB14}. 
The networks and sub-networks were initialized using the parameters of 
$\tilde{\times}$ presented in \cite{deep_learning_schwab} for the first example 
and $\tilde{f}$ for the rest of the examples using a practical network architecture. 
The polynomial coefficients used in the first example were computed using 
built-in \texttt{NumPy} functions.

\subsection*{Training a Network From a Polynomial 
Initialized State}

\begin{figure}
        \centering
        \begin{subfigure}{.45\textwidth}
            \centering
            \includegraphics[width=\textwidth]{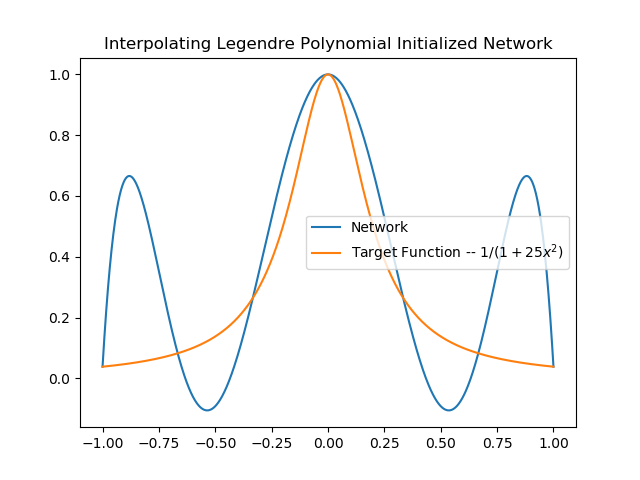}
            \caption{}
            \label{fig:polyinit_rational}
        \end{subfigure}%
        \begin{subfigure}{.45\textwidth}
          \centering
            \includegraphics[width=\textwidth]{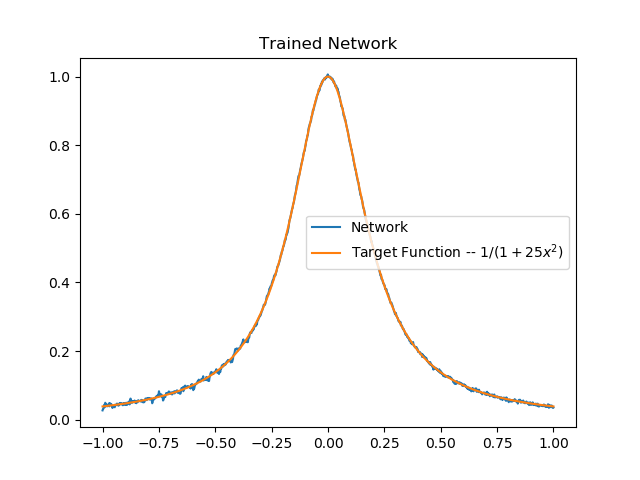}
            \caption{}
            \label{fig:trained_rational}
        \end{subfigure}
        \caption{ We show that an approximation of a rational polynomial can be learned 
        from uniformly spaced
        samples. Here we plot: (a) A network initialized to a degree $6$ 
            Legnedre polynomial interpolation of a rational function, and
            (b) The behavior of the polynomial initialized network after training.}
    \label{fig:rational_poly_poly_approx}
    \end{figure}

We consider learning the rational polynomial
$R(x) = \frac{1}{1+25x^2}$ from a set of equally spaced 
samples from the interval $[-1,1]$. A network which has been initialized
to approximate the degree $6$ Legendre interpolant of $R$
is shown in Figure \ref{fig:polyinit_rational}.
From this initialized state, we train all network parameters. The result of the trained network is shown
in Figure \ref{fig:trained_rational}. Notice that the trained network is a better 
approximation to the target than the polynomial used to initialize it.
    
    \subsection*{A Variation on the Training Procedure}
\begin{figure}
        \begin{subfigure}{.33\textwidth}
            \centering
            \includegraphics[width=\textwidth]{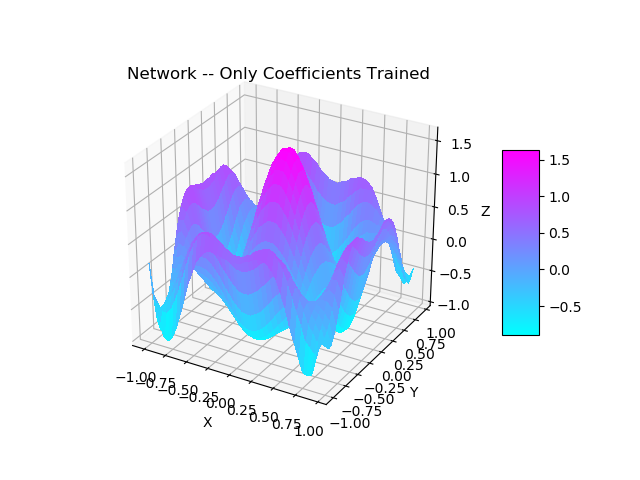}
            \caption{}
            \label{fig:polyinit_cos2}
        \end{subfigure}%
        \begin{subfigure}{.33\textwidth}
          \centering
            \includegraphics[width=\textwidth]{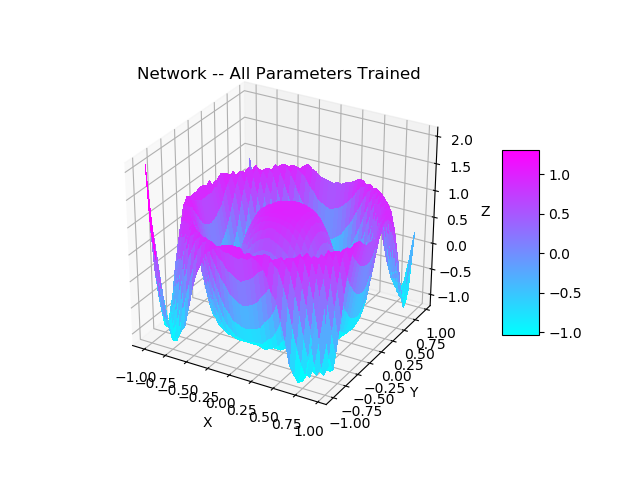}
            \caption{}
            \label{fig:trained_cos2}
        \end{subfigure}%
        \begin{subfigure}{.33\textwidth}
          \centering
            \includegraphics[width=\textwidth]{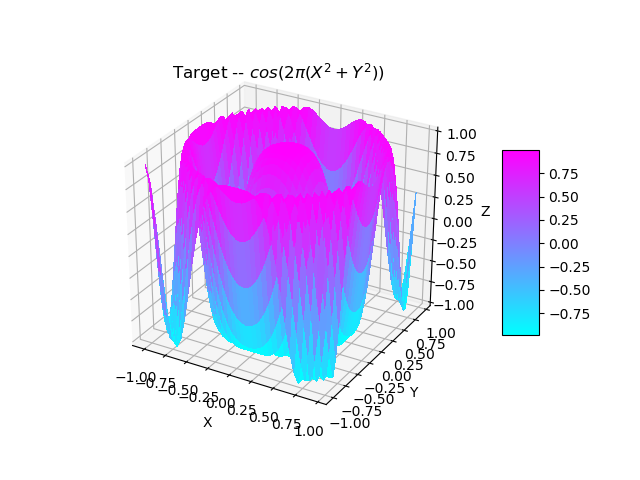}
            \caption{}
            \label{fig:target_cos2}
        \end{subfigure}
        \caption{We initialize the network so that the product blocks $\tilde{\Psi}$ compute 
        the products associated with tensor products of the Legendre Polynomials but do not 
        set the Legendre coefficients as the output weights. Here we plot
        (a) the approximating network after polynomial initialization and after
        training only the polynomial coefficients,
        (b) the network after training all parameters, and
        (c) the target function $T$. }
    \label{fig:cos2_poly_approx}
    \end{figure}

Next, we consider approximating the function 
$T(x_1,x_2) = \cos (2\pi(x_1^2+x_2^2))$, displayed in 
Figure \ref{fig:cos2_poly_approx} (c),
from uniform random samples in $[-1,1]^d$. 
In the previous example we computed the polynomial
coefficients with respect to some interpolating points. However, one can 
also consider learning these coefficients. In this example, we
initialize a network with respect to 
a polynomial in $2$ dimensions associated with the total degree space of 
Legendre polynomials of order $8$, i.e.,
    \begin{equation}
        P(x_1,x_2) = \sum_{|i+j|\le 8} c_{i,j} L_i(x_1) L_j(x_2),
    \end{equation}
where $L_k$ is the $k^{th}$ degree Legendre polynomial.
Once the all network parameters, except for the output weights,  
were initialize, we trained only the weights $c_{i,j}$.
The result of this training procedure is shown in Figure \ref{fig:cos2_poly_approx} (a).
Having reached a reasonable polynomial approximation, we then trained all network 
parameters. The fully trained network is plotted in Figure \ref{fig:trained_cos2}.
Similarly to the previous example, the trained network performs better than 
the polynomial approximation used to initialize it.

\subsection*{A Practical Network based on Polynomials}
The network $\tilde{F}$ can grow very large in terms of the total number
of trainable parameters if $\sum_{i=1}^N \|\vec{\nu}_i\|_1$ is large because
high-degree polynomials require many multiplications. 
Since our main motivation is not to produce a network that behaves like a
polynomial, but rather to produce a network with suitable architecture and
initialization so that it can learn complicated functions effectively,
it is reasonable to construct and initialize a network with 
a low degree polynomial such as $\tilde{S}$ (\ref{eq:proposed_net}). 
This network has a commonly used network architecture and is 
implemented as a series of linear layers with ReLU activation between. 

\begin{figure}
\begin{tabular}{ccc}
  \includegraphics[width=45mm]{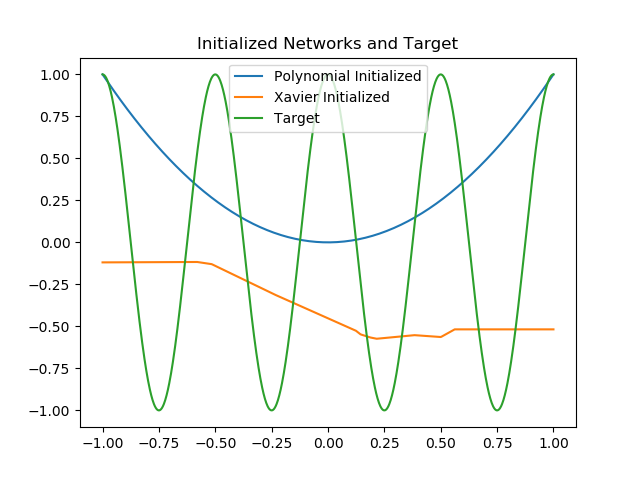} &  
  \includegraphics[width=45mm]{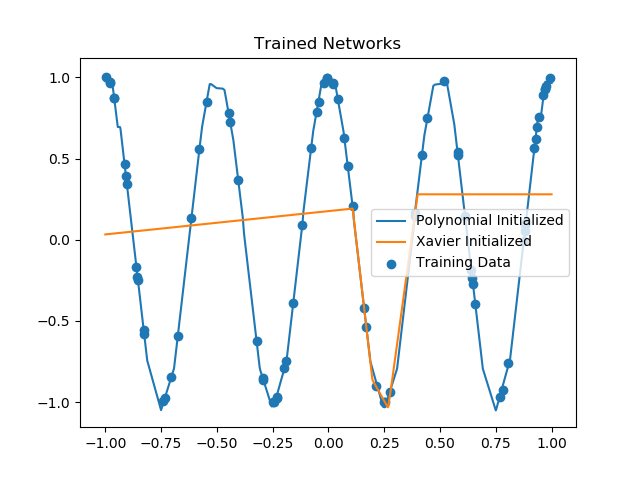} &
  \includegraphics[width=45mm]{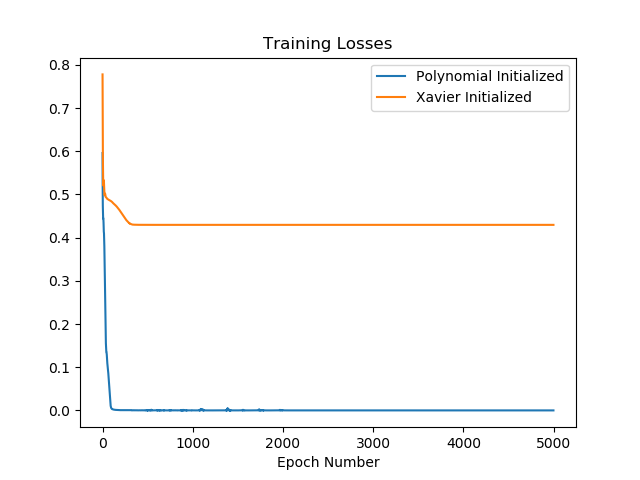} \\
  (a) & (b) & (c) \\[6pt]
\end{tabular}
\caption{We compare training our network for $x^2$, with $6$ hidden layers and $4$ nodes per 
layer, to approximate the function $\cos(4\pi x)$ on the interval [-1,1] from a polynomial initialized state
and from a randomly initialized state. Here we plot: (a) The target function, the polynomial initialized network and
the randomly initialized network; (b) the network obtained after training starting from both a polynomial and
random initial states; and (c) the training losses associated with both initialization procedures.
}
    \label{deep_rational}
\end{figure}

In Figure \ref{deep_rational} we consider approximating the function 
$\cos(4\pi x)$ on the interval [-1,1] using the network $\tilde{f}$
and using $80$ randomly chosen sample points
as training data. We compare training this network from a polynomial
initialized state, i.e., initialized to approximate $x^2$ 
and a randomly initialized state chosen via the method proposed in 
\cite{glorot2010understanding} sometimes called Xavier initialization. 
The initialized networks are depicted in
\ref{deep_rational} (a). The trained networks for the polynomial initialized 
and randomly initialized cases are plotted in 
\ref{deep_rational} (b). 
Notice that the polynomial initialized network performs better for points that were not sampled. 
Moreover, according to the training losses in \ref{deep_rational} (c), the polynomial
initialized network learned parameters associated to a more desirable local minimum
more quickly than the randomly initialized network.

Although deep network are known to be more expressive than shallow ones \cite{telgarksy_benefit_depth}, 
they have a tendency to over-fit the training data \cite{NIPS2012_4824}. One way to measure over-fitting of the
data is to compare the error on a small training set to the error on a large validation set, i.e., a set of samples
of the target function not used for training. We now show that our polynomial initialized network can efficiently 
learn high-dimensional functions from a relatively small training set. 

\begin{figure}
    \centering
\begin{tabular}{cc}
  \includegraphics[width=55mm]{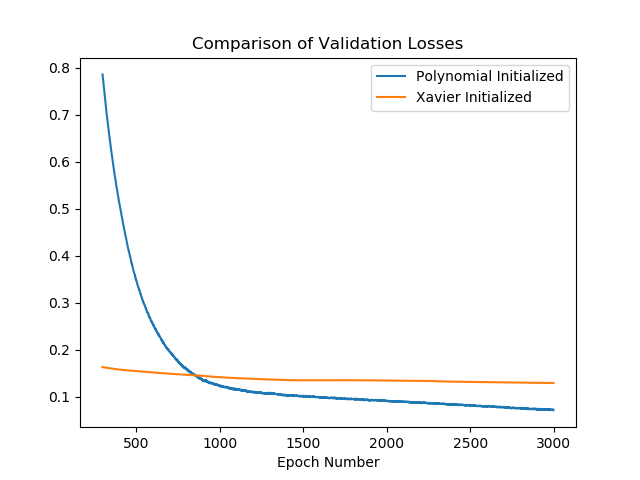} &  
  \includegraphics[width=55mm]{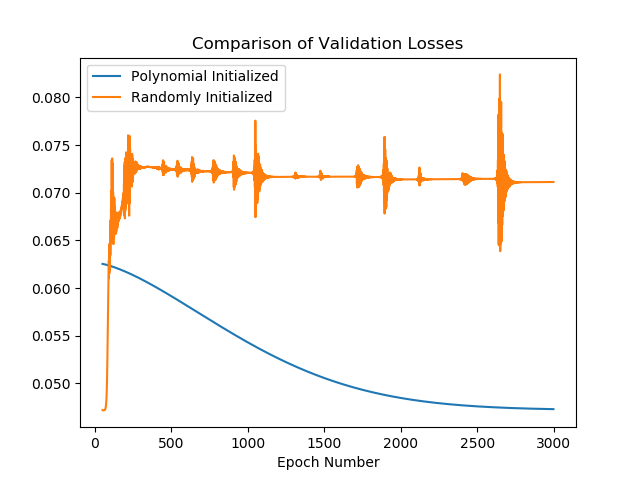}  \\
  (a) & (b) \\[6pt]
\end{tabular}
\caption{Our polynomial initialization helps prevent over-fitting of the training data 
during the training of a deep network for approximating the function $G$ 
with $a_i = 5$ and $u_i = 1/2$. We plot: 
(a) The validation losses for approximation of $G$ with $d=4$; and 
(b) The validation losses for approximation of $G$ with $d=20$.}
    \label{Gapprox}
\end{figure}

Let $G$ be the $d$-dimensional function 
    \[
        G(\vec{x}) = \exp \left ( - \sum_{i=1}^d a_i | x_i - u_i| \right )
    \]
which is used to test high-dimensional integration routines \cite{genz}. We will compare
training two networks which the same architecture. One will be randomly initialized
using the built-in random initializer of \texttt{PyTorch} and the other will be initialized to
the polynomial associated with $\tilde{S}$ from (\ref{eq:proposed_net}). 
In Figure \ref{Gapprox} (a) we we consider approximating $G$ where $d=4$ with a deep network with
$20$ Layers and $28$ nodes per layer from $500$ uniform random samples
on $[0,1]^d$. The validation set is composed of $3000$ uniform random points.
While both networks have decreasing validation loss and therefore are approximating the
function well outside of the training set, the polynomial initialized network was able to achieve 
better performance. In Figure \ref{Gapprox} (b) we compare the result of approximating $G$ for $d=20$ 
by two networks, each with $8$ Layers and $156$ nodes
on each layer, but one with polynomial initialization and the other Xavier initialization. 
The training set is made up of $300$ uniform random samples on $[0,1]^d$ and
the validation set is made up of $5000$ uniform random samples on $[0,1]^d$.
From the validation losses plotted in Figure \ref{Gapprox} we see that our polynomial
initialization allows for learning which decreases the validation error and hence
is not as affected by over-fitting phenomenon common to deep networks. 

\section{Conclusion}
The connection established in this effort
between polynomials and networks gives a heuristic for choosing 
suitable network hyper-parameters.
As shown in our numerical examples, our presented networks may be able to 
find better local minima of the loss landscape through
their connection to polynomial approximation which determines an initialization
of parameters as well as an architecture. The numerical examples 
focused on functions which are real valued, but extending this work
to functions from $\mathbb{R}^d \rightarrow \mathbb{R}^k$ can be accomplished
by approximating each component of the output with the same set of polynomials.
In future work, we plan to apply our initialization to networks used
in high-dimensional classification problems.
Another possible extension of this work is to explicitly construct networks
that achieve other kinds of classical approximations, such as in an arbitrary 
orthogonal basis. In addition, we consider only ``global" polynomial approximation, 
i.e., polynomials with support on the entire interval on which we hope to approximate a target. 
It would be interesting to consider how a network could be constructed which approximates a 
piecewise polynomial function or some other function which can be
expressed by local basis. 

\subsubsection*{Acknowledgments} 
This material is based upon work supported in part by: the U.S. Department of Energy, Office of Science,
Early Career Research Program under award number ERKJ314; 
U.S. Department of Energy, Office of Advanced Scientific Computing Research under award numbers 
ERKJ331 and ERKJ345; the National Science Foundation, Division of Mathematical Sciences, 
Computational Mathematics program under contract number DMS1620280; and by the Laboratory Directed 
Research and Development program at the Oak Ridge National Laboratory, which is operated by UT-Battelle, LLC., 
for the U.S. Department of Energy under contract DE-AC05-00OR22725.

We would like to acknowledge Hoang A. Tran for many helpful discussions.

\bibliography{dnn_references}

\end{document}